%% file: egpaper_final.tex
\documentclass[10pt,twocolumn,letterpaper]{article}

\usepackage{iccv}
\usepackage{times}
\usepackage{epsfig}
\usepackage{graphicx}
\usepackage{amsmath}
\usepackage{amssymb}
\usepackage{adjustbox}
\usepackage{float}
\usepackage{makecell}
\usepackage{xcolor}

\usepackage[numbers]{natbib}
\usepackage{graphicx}
\usepackage{amssymb}
\usepackage{amsmath}
\usepackage{pifont}

\usepackage{scalerel,stackengine}
\stackMath
\newcommand\reallywidehat[1]{%
\savestack{\tmpbox}{\stretchto{%
  \scaleto{%
    \scalerel*[\widthof{\ensuremath{#1}}]{\kern.1pt\mathchar"0362\kern.1pt}%
    {\rule{0ex}{\textheight}}
  }{\textheight}%
}{2.4ex}}%
\stackon[-6.9pt]{#1}{\tmpbox}%
}

\newcommand*{\vertbar}{\rule[-1ex]{0.5pt}{2.5ex}}

\usepackage[pagebackref=true,breaklinks=true,letterpaper=true,colorlinks,bookmarks=false]{hyperref}

\def\iccvPaperID{915}

\iccvfinalcopy 

\ificcvfinal\fi
\begin{document}

\title{Learning a Disentangled Embedding for Monocular 3D Shape Retrieval and Pose Estimation}

\author{Kyaw Zaw Lin\\
National University of Singapore\\
{\tt\small kyawzl@comp.nus.edu.sg}
\and
Weipeng Xu\\
Max Planck Institute for Informatics \\
{\tt\small wxu@mpi-inf.mpg.de}
\and
Qianru Sun\\
Max Planck Institute for Informatics \\
{\tt\small qsun@mpi-inf.mpg.de}
\and
Christian Theobalt\\
Max Planck Institute for Informatics \\
{\tt\small theobalt@mpi-inf.mpg.de}
\and
Tat-Seng Chua\\
National University of Singapore \\
{\tt\small dcscts@nus.edu.sg}
}

\author{Kyaw Zaw Lin$^{1}$ \quad Weipeng Xu$^{2}$ \quad Qianru Sun$^{1,2}$ \quad Christian Theobalt$^{2}$  \quad Tat-Seng Chua$^{1}$\\
\\
\small  $^{1}$National University of Singapore  \\
\small  $^{2}$Max Planck Institute for Informatics, Saarland Informatics Campus\\
\small {\texttt{kyawzl@comp.nus.edu.sg}}  {\texttt{\{wxu,qsun, theobalt\}@mpi-inf.mpg.de}} \\
\small  \quad  {\texttt{\{dcssq, dcscts\}@nus.edu.sg}}
}

\maketitle

\input{sec/abstract.tex}

\input{sec/intro.tex}
\input{sec/related.tex}
\input{sec/method.tex}
\input{sec/exp.tex}
\input{sec/conclusion.tex}

\section{Acknowledgement}
This  research  is  part  of  NExT  research  which is  supported  by  the  National  Research  Foundation,  Prime  Minister’s  Office,  Singapore  under  its IRC@SG Funding Initiative.   It is partially supported by German Research Foundation (DFG CRC 1223).

{
\bibliographystyle{ieee}
\bibliography{egbib}
}

\end{document}

%% file: sec/abstract.tex
\begin{abstract}
We propose a novel approach to jointly perform 3D shape retrieval and pose estimation from monocular images.
In order to make the method robust to real-world image variations, e.g. complex textures and backgrounds, we learn an embedding space from 3D data that only includes the relevant information, namely the shape and pose. 
Our approach explicitly disentangles a shape vector and a pose vector, which alleviates both pose bias for 3D shape retrieval and categorical bias for pose estimation.
We then train a CNN to map the images to this embedding space, and then retrieve the closest 3D shape from the database and estimate the 6D pose of the object.
Our method achieves 10.3 median error for pose estimation and 0.592 top-1-accuracy for category agnostic 3D object retrieval on the Pascal3D+ dataset, outperforming the previous state-of-the-art methods on both tasks.

\end{abstract}

%% file: sec/intro.tex
\section{Introduction}
\begin{figure}[t]
 \centering
   \includegraphics[width=0.8\linewidth]{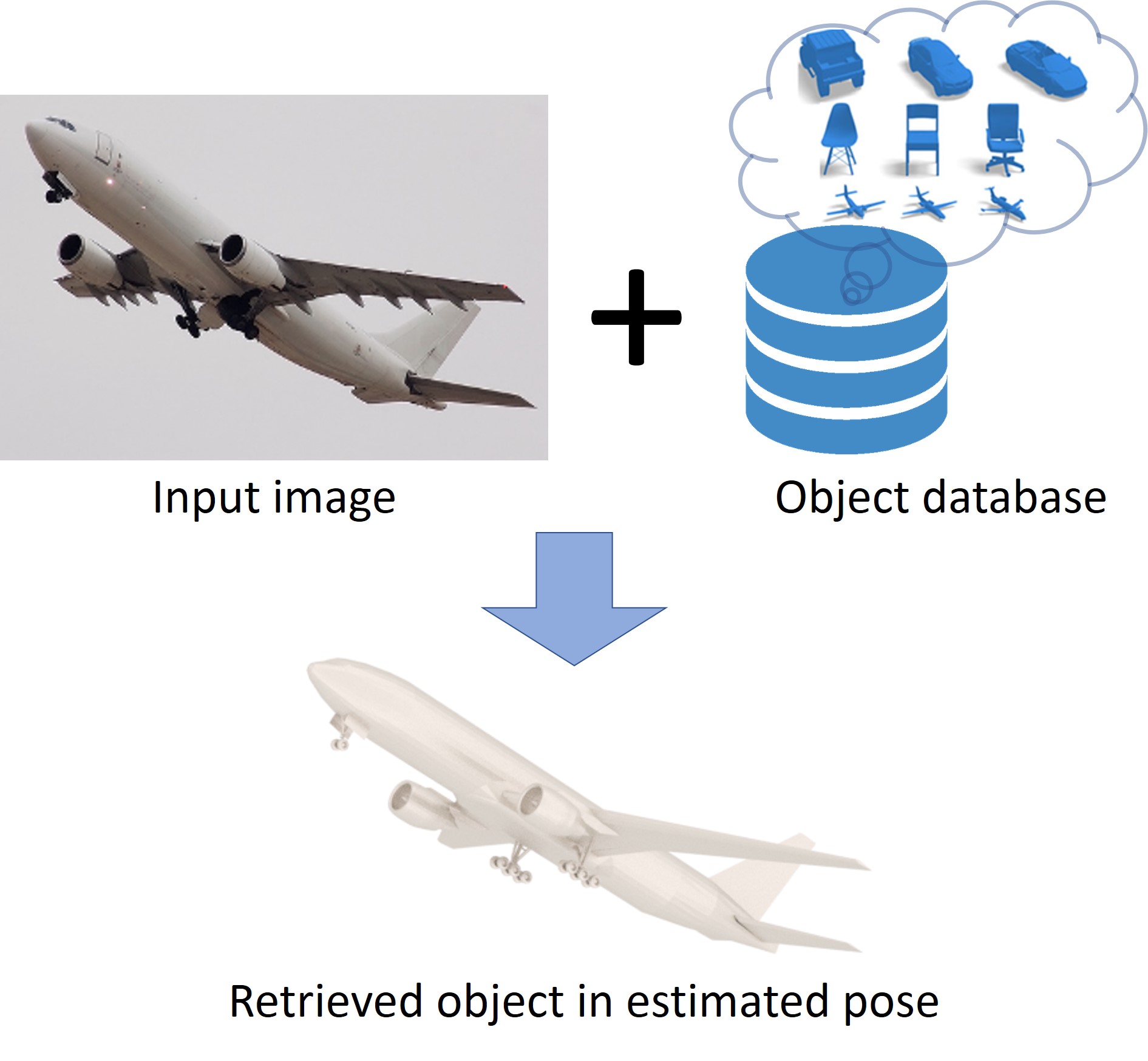}
   \caption{Given a single RGB image of an object, our method retrieves its closest 3D shape from a database and estimates its 6D pose.}
\label{fig:task}
\end{figure}

The tasks of estimating 3D shape and pose from monocular images (see Figure.~\ref{fig:task}) are highly correlated and under-constrained.
Solving these two problems jointly is an important research direction, which has a broad range of applications in many areas such as augmented reality, 3D scene understanding and robotics.
For the 3D shape retrieval task, an essential problem is how to deal with the pose variance of an object.
To this end, many of the existing methods learn a pose-invariant embedding, using multi-view rendered images of CAD models in different poses~\cite{aubry2014seeing,aubry2015understanding,izadinia2017im2cad,grabner20183d}.
However, there is a significant gap between rendered images and real images due to distracting factors such as varying lighting conditions, camera response functions and backgrounds.
Moreover, one has to render a large number of images from CAD models with different textures and poses in order to cover the appearance variation.
However, simulating all possible variations graphically is prohibitive and still generalizes poorly when applied to natural images.

In order to alleviate this problem, \cite{grabner20183d} proposes 1) to render depth images instead of RGB images to circumvent the real~vs.~synthetic domain gap due to texture, lighting and background, and 2) to first estimate the pose of the object and only render the 3D models in the estimated pose, which significantly reduces the rendering cost during testing.
However, they still have to render the 3D models in densely sampled poses for training, and more importantly, errors in the pose estimation lead to failures in 3D shape retrieval.

In this paper, we propose a novel approach to joint 3D shape retrieval and pose estimation based on learning a disentangled shape and pose embedding.
Our method consists of two stages:
In the first stage, we train a CNN to learn an embedding space from 3D data, which only encodes the relevant scene independent information, namely shape and pose, and therefore is free of environment dependent factors.
Comparing to learning from images, this allows us to eliminate the distracting factors such as texture, lighting and background, because they do not exist in the 3D data.
Furthermore, we explicitly disentangle the shape and pose embedding.
To this end, we train the network with 3D volumes of occupancy grids of the objects in different poses and map them to a pair of embedding vectors, i.e. a shape vector and a pose vector.
In the second stage, we train another CNN to map the 2D image to the embedding vectors, which allows us to retrieve the 3D shape and estimate its 3D pose simultaneously from the image. 

Benefiting from the disentanglement of shape and pose, the learned shape embedding is invariant to the pose and the pose embedding is invariant to the shape.
The disentanglement alleviates both the categorical bias for pose estimation and pose bias for shape retrieval.
On the one hand, our \textit{shape retrieval} does not rely on the pose estimation and outperforms the existing methods by a large margin.
On the other hand, we achieve state-of-the-art \textit{pose estimation} performance using just a single category-agnostic network, which is in contrast to many existing methods that utilize one sub-network per category to exploit category dependent pose distributions.
Furthermore, our approach does not rely on generating multi-view rendered images with pose variations, which is difficult in practice.
Instead, our volumetric training samples in different poses are generated efficiently in an online manner.

In summary, our contributions are as follow:
\begin{itemize}
    \item We propose to learn the disentangling of the shape and the pose embedding spaces, which allows us to remove categorical bias in both pose and shape representation and therefore improves the performance of both tasks.
    \item We propose a new pose estimation method that extends the 6D-rotation-representation-based method of~\cite{zhou2018continuity} to the hybrid classification-regression strategy.
    \item We show that our proposed approach outperforms the state-of-the-art methods on the Pascal3D+ dataset~\cite{xiang2014beyond} on both 3D shape retrieval and pose estimation tasks.
\end{itemize}

%% file: sec/related.tex
\section{Related work}
There is a rich literature on image based object retrieval and pose estimation.
We will limit the discussion in this section to the topics that are most related to our work: image-based 3D shape retrieval, 3D pose estimation and joint representation learning.

\noindent\textbf{Image-based 3D Shape Retrieval.} Methods for 3D shape retrieval rely on synthesized images of 3D models to perform matching between images and 3D models \cite{aubry2014seeing,aubry2015understanding,izadinia2017im2cad,grabner20183d}.
However, it is challenging to perform direct comparison between CNN features of rendered images and real images due to the domain gap. 
To circumvent the problem arising from the use of rendered images, many have attempted to use the intrinsic representation of 3D models directly instead of rendered images to represent 3D models. %
Li \etal ~\cite{li2015joint} mapped image features to light field descriptors computed from 3D models.  
Tasse  \etal~\cite{tasse2016shape2vec} used the word2vec \cite{goldberg2014word2vec} as an embedding space and map different input modalities such as 3D model, image and sketch drawing to the same space. 
Girdhar  \etal~\cite{girdhar2016learning} used voxel reconstruction as embedding space but did not consider pose variations.

Our proposed method has several advantages over existing works.
We consider fine-grained instance level retrieval, unlike \cite{li2015joint,tasse2016shape2vec}, where they only consider category-level retrieval.
Grabner \etal~\cite{grabner20183d} need to render 3D model in estimated viewpoint and search among multiple models, as they does not explicitly decouple pose in their representation.
This leads to failure in 3D shape retrieval when pose estimation fails.
In our method, our shape embedding is invariant to input's pose and vice versa.
This allows us to simply match against shape embeddings from any viewpoints, without having to take into consideration variations in shape embedding due to viewpoint differences. 
Unlike \cite{li2015joint,girdhar2016learning}, where only the shape is encoded in the embedding, our disentangled embeddings encode both shape and pose.

\noindent\textbf{3D Pose Estimation.} There are two main categories of approaches to 3D pose estimation.
The first category is based on keypoints detection.
These approaches assume that the 3D model of the object is available and predict 2D keypoints with known correspondences on the 3D model. 
Then a perspective-n-point problem (PnP) is solved to find the transformation parameters, which minimizes the distance from 2D projection of 3D points to detected 2D keypoints \cite{Jourabloo_2016_CVPR,kumar2017kepler,zhu2016face}.

This technique is commonly used on the specific object classes with limited variations across object instances, for example, on faces.
To make this technique applicable to general objects, the objects are typically approximated by their 3D bounding boxes and the 8 or 9 points of the enclosing cubes are predicted \cite{rad2017bb8,grabner20183d}.
However, it it necessary to be provided ground truth 3D dimensions \cite{rad2017bb8} or to learn to predict the 3D dimension of the box in addition to the keypoints \cite{grabner20183d}.
In our approach, we directly output a rotation matrix and requires no additional information, which is more scalable across more object categories.

The second category of approaches directly predict the transformation parameters. 
Su \etal ~\cite{tulsiani2015viewpoints} represent rotations as bins in Euler angles and formulate pose estimation as a classification problem. 
However, the quantization of angles introduces inaccuracy, even though the classification predictions are accurate. 
Su \etal~\cite{su2015render} address the problem using a geometry aware soft weighted classification scheme.
Mahendran \etal~\cite{mahendran20173d} attempt to improve on it by treating it as a regression problem, which is a more natural formulation of the task.
They represent rotations in axis-angle or quaternion space and use geodesic distance as an alternative to L2 distance.
However, their method cannot outperform the classification approach in \cite{su2015render}.

Recently, hybrid approaches based on classification followed by residual regression \cite{ren2015faster,guler2017densereg,rogez2017lcr} have become popular for a variety of different tasks.
Such approaches have been applied to pose estimation to achieve state-of-the-art performance on Pascal3D+ dataset \cite{mahendran2018mixed,li2018unified}.
We adopt this hybrid strategy in our framework for pose estimation.
Comparing to existing methods, our approach achieves additional robustness afforded by the guidance from the ``pure'' information learned from 3D data, which is free from distracting factors in the images.

\noindent\textbf{Joint Representation Learning.} The idea of using a common and meaningful embedding space has a rich history.
A typical use is to construct joint image-to-text embeddings \cite{gong2014multi,frome2013Devise}, where images and text can be compared directly, facilitating text to image retrieval or vice versa. 
Following the same idea, common embedding space for shape was proposed in \cite{tasse2016shape2vec}, where multiple modalities (3D mesh, image, text etc.) are embedded using the word2vec semantic space. 
Different input modalities for embeddings have been investigated such as shapes and text \cite{chen2018text2shape},  shapes and images \cite{wu2018joint}, and sketches and 3D shapes \cite{dai2018deep}. 
Majority of the existing works are concerned with cross modal retrieval and did not consider such embedding spaces to be an effective proxy for improving task specific performance.
In our proposed method, we learn a task specific embedding space shared by images and 3D shapes, which encodes the shape and pose information in a disentangled way.
Our experiments show that such a disentangled embedding space is effective for improving the performance of both tasks.

%% file: sec/method.tex
\section{Method}

\begin{figure}[ht]
    \centering
   \includegraphics[width=\columnwidth]{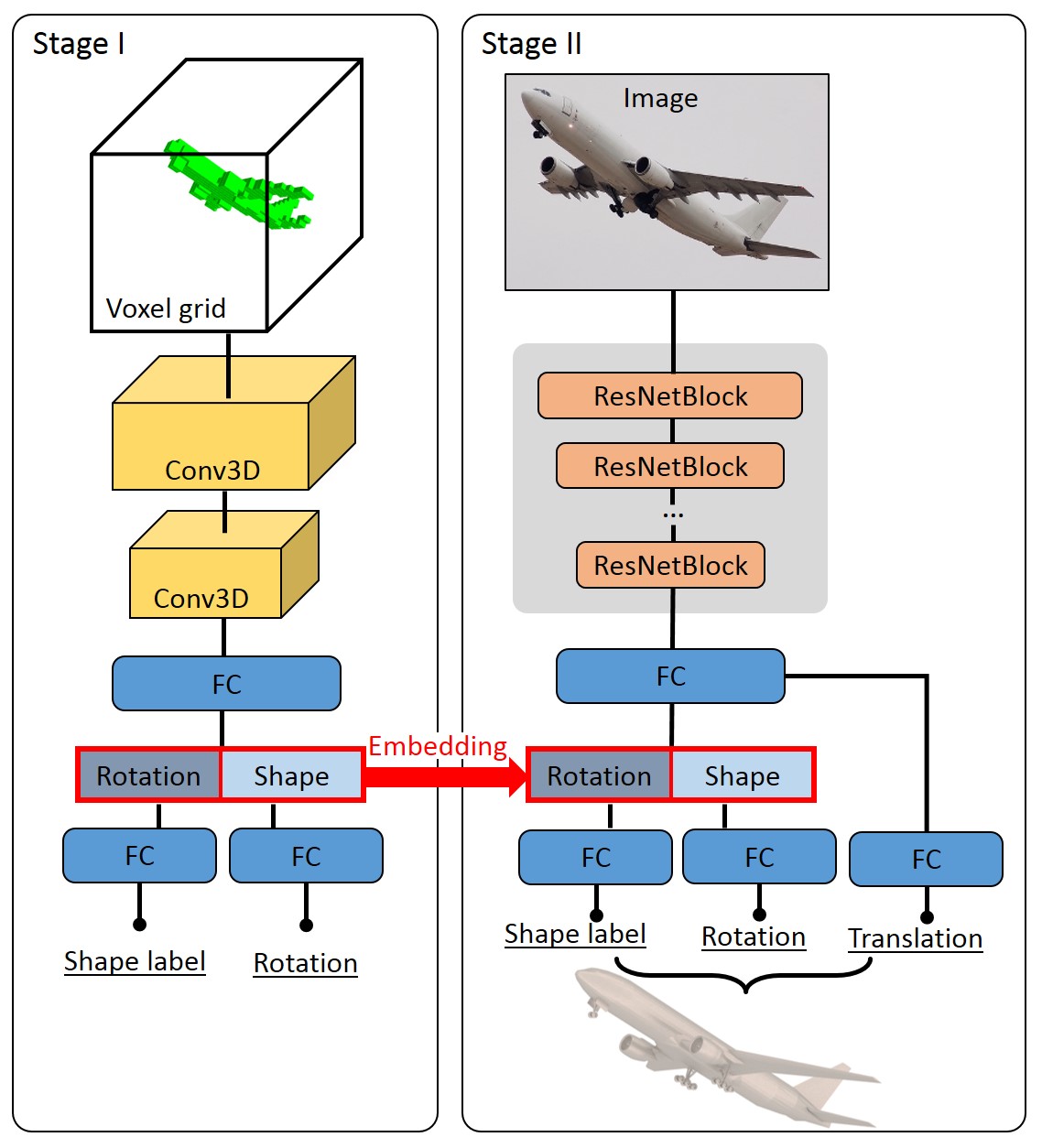}
   \caption{
   Our method consists of two stages: First, we learn disentangled embedding space from 3D data, which is free of distracting factors. Then, we train a CNN to map the 2D images to the embedding vectors, which allows us to retrieve the 3D shape and estimate the pose simultaneously for the images. 
   }
\label{fig:framework}
\end{figure}

As shown in Figure~\ref{fig:framework}, our method consists of two stages.
In the first stage, we train a CNN (referred as \textit{embedding network}) to learn a disentangled embedding space for shape and pose. 
In the second stage, we train another CNN (referred as \textit{regression network}) to map a monocular 2D image to the embedding vector, which allows us to retrieve the 3D shape and estimate the 3D pose simultaneously from the images.

\subsection{ \mbox{Learning the Embedding from 3D Data (Stage I)}}
\noindent\textbf{Voxel-based Learning of Embeddings.}  Our disentangled shape and pose embedding is illustrated in Figure~\ref{fig:framework} \textbf{Stage I}.
We use the 3D volumetric occupancy grid, which is a natural representation for 3D objects, as the input, and learn a pure shape and pose embedding from it.
The input 3D volumetric occupancy grid is generated with the \textit{binvox}~\cite{binvox} tool using the CAD models of the objects.
The resolution of the input volume is $32\times 32\times 32$.
We map the input volume to the concatenated embedding vectors of shape and pose using an architecture similar to \textit{VoxNet}~\cite{maturana2015voxnet}, which consists of two 3D convolution layers and one fully connected layer.

\noindent\textbf{Shape Embedding.} For training the shape embedding, we impose a cross entropy loss for shape classification.
Note that the cross entropy loss can be replaced with a pair-wise distance metric such as triplet ranking loss~\cite{schroff2015facenet}, when dealing with a dataset that consists of a large amount of objects, e.g. the ObjectNet3D dataset~\cite{xiang2016objectnet3d}.
In our experiments, we use the Pascal3D+ dataset~\cite{xiang2014beyond} that contains 79 objects, where the cross entropy loss is more efficient for training.

\noindent\textbf{Pose Embedding.} In order to learn the pose embedding, we apply random rotations on the 3D volumes during training.
Note that the translation is not encoded in the pose embedding, since we observe that the translation is dependent on the image cropping.
Therefore, we regress the translation from the input image directly in the second stage.

\noindent\textbf{Pose Estimation.} 
We formulate pose estimation task as hybrid classification-regression problem, which is also called bin and delta technique in the literature \cite{li2018unified,mahendran2018mixed}.
We first solve the classification problem by predicting a coarsely discretized rotation bin, followed by a fine-grained regression step which predicts a continuous delta rotation within the bin.
The predicted rotation $R$ is obtained via the equation
\begin{equation}
\label{eq:rot_pred}
R = \hat{R}^TR_{d},
\end{equation}

where $\hat{R}$ is predicted discrete rotation bin obtained from classification and $R_{d}$ is the regressed delta rotation.
The set of rotation bins $\{R_{i}\}_{i=1,...,N}$ is obtained by discretizing $\mathcal{SO}(3)$ into $n$ equal bins using the method and software provided by \cite{yershova2010generating}.
We then minimize the binary cross entropy loss.
\begin{equation}
\label{eq:rot_bin}
\mathcal{L}_{bin\_r}(x,y) = \dfrac{1}{N} \sum\limits_{i}^{N} y_i~log(x_i)+(1-y_i)~log(1-x_i)
\end{equation}
Where $x$ and $y$ are probability of the predicted rotation bin and the ground truth respectively.
$N$ is the number of rotation bins.
We use a soft labeling scheme for $y$ where

\begin{equation}
\label{eq:soft_rs}
y_i= \begin{cases}
1 & if~ GD(R_i,R_{gt})=0  \\
\alpha & if~ GD(R_i,R_{gt})<\beta \\
0 \\
\end{cases}
\end{equation}

$GD(R_1,R_2)$ is the geodesic distance which can be computed as:
\begin{equation}
\label{eq:gd_v1}
GD(R_1,R_2) = cos^{-1}\left( \dfrac{ tr(R_1 R_2^T)-1}{2}\right)
\end{equation}

Our delta rotations are formulated as 6-dimensional representation, which avoids the continuity problems present in other rotation representations such as Euler angles, axis-angles or quaternions, as shown by \cite{zhou2018continuity}.
Specifically, we predict a 6-dimensional representation $\{a_1,a_2\} \in \mathbb{R}^{3\times2}$ which is transformed into an orthogonal rotation matrix $R$ by the following procedure:

\begin{equation}
f\left(\left[ {\begin{array}{ccc}
   \vertbar & \vertbar \\
   a_1      & a_2 \\
   \vertbar & \vertbar
  \end{array} } \right]\right) = 
  \left[ {\begin{array}{ccc}
   \vertbar & \vertbar & \vertbar \\
   b_1 & b_2 & b_3 \\
   \vertbar & \vertbar & \vertbar
  \end{array} } \right]
\end{equation}
\begin{equation}
b_i = \left[ \begin{cases}
N(a_1) & \text{if } $i=1$ \\
N(a_2 - (b_{1} \boldsymbol{\cdot} a_2) b_1) & \text{if } $i=2$ \\
b_1 \times b_2 & \text{if } $i=3$
\end{cases}
\right]
\end{equation}
\begin{equation}
N(a) =  \dfrac{a}{\|a\|}
\end{equation}

Delta rotations are learnt by minimizing the geodesic distance of all the bins activated by Eq. \ref{eq:soft_rs} and their corresponding rotations, written as:

\begin{equation}
\label{eq:rot_delta}
\mathcal{L}_{delta\_r} = \dfrac{1}{N} \sum\limits_{i}^{N} GD(\hat{R}_{di},R_{di}) 
\end{equation}
%
We note that this soft labelling scheme is reminiscent of the geometric structure aware loss function introduced by \cite{su2015render}, which applies the soft labels on Euler angles.
However, ours is more geometrically meaningful due to the uniform quantization of the rotation group and the use of geodesic distance as a metric to measure closeness of rotations.

\subsection{3D Shape Retrieval and Pose Estimation from Images (Stage II)}
In the second stage of our method, we retrieve the 3D shape and estimate the 6D pose of the object.
As shown in Figure~\ref{fig:framework} \textbf{Stage II}, we train a ResNet~\cite{he2016deep} to map the image to the ground truth embedding vectors with the L1 loss, $\mathcal{L}_{embed}$ .
The ground truth embedding vectors are obtained by applying the \textit{embedding network} on the 3D volume of the ground truth shape in the ground truth rotation.
In addition, similarly to the \textit{embedding network}, the \textit{regression network} contains a sub-network for shape classification trained with a cross-entropy loss, $\mathcal{L}_{cls}$ and another sub-network for the rotation regression trained with an bin and delta loss similar to Eq. \ref{eq:rot_bin} and \ref{eq:rot_delta}.

The absolute 3D translation of the object cannot be obtained without knowing the camera intrinsic parameters, the image cropping and the dimension of the object.
However, in order to be able to overlay the object onto the image, we estimate the up-to-scale 3D translation with respect to the image cropped with the object bounding box, assuming a common camera model and normalized 3D shape.
Regression of the 3D translation is performed similarly as the rotation regression with a sub-network trained using bin and delta loss.
To this end, the translation bins are obtained by dividing the Euclidean space into equal cubes centered at $\mathbf{t}_{bin}$.
During training, we normalize the residual translation $\mathbf{t}_{delta}$ using the dimension of the corresponding bin cube to constrain it within range $[0,1]$. 
We use cross entropy loss for the bin classification, $\mathcal{L}_{bin\_t}$  and Huber loss for regressing the translation residual within the bin, $\mathcal{L}_{delta\_t}$ 
The overall loss function for training the \textit{regression network} is thus:

\begin{equation}
\resizebox{.85\hsize}{!}{
$\mathcal{L} = \mathcal{L}_{embed} +  \mathcal{L}_{cls} +  \mathcal{L}_{bin\_r} +  \mathcal{L}_{delta\_r} + \mathcal{L}_{bin\_t} +  \mathcal{L}_{delta\_t}$}
\end{equation}

During testing, we only feed a single image into the \textit{regression network} and find the bin with the highest score and apply the corresponding deltas to obtain 6-DOF pose estimation prediction. For retrieval, we obtain the shape label by finding the closest 3D model embedding via L2 distance.

\begin{equation}
\hat{i} = \operatorname*{argmin}_i \lVert f(x_i) - x  \rVert ,
\end{equation}
where $f(x_i)$ is the 3D shape embedding in canonical pose obtained from the \textit{embedding network} and $x$ is the image-based shape embedding from the \textit{regression network}.
Note that we can choose to obtain $f(x_i)$ under any pose as our shape embeddings are disentangled from pose.

\subsection{Implementation Details}
For training the \textit{embedding network} in Stage I, we use a similar architecture as \textit{VoxNet} \cite{maturana2015voxnet}.
The dimensionality of the embedding vectors for both shape and pose is 512. 
For our \textit{regression network} in Stage II, we use the ResNet50~\cite{he2016deep} pre-trained on ImageNet~\cite{deng2009imagenet} as the backbone network.
Our networks are trained with Adam optimizer with a learning rate of $10^{-4}$.
We train the \textit{embedding network} for 100 epochs and we generate 10,000 samples during each epoch. 
Our \textit{regression network} is also trained for 100 epochs, where each epoch constitutes all the training examples in the dataset.
We also apply small jittering of 5\% of the object's dimension and blurring with a probability of 20\%. 
For training the \textit{regression network}, the range of the translation vector is restricted to $x\in[-0.25,1.5]$, $y\in[-0.25,1.5]$ and $z\in[0.5,10.0]$, which is determined by the translation range in the training data.
More detailed information including network architecture can be found in our supplementary material.

\noindent \textbf{Dataset}. We used the Pascal3D+ \cite{xiang2014beyond} dataset for our training data. In this dataset, there are 13,898 object instances that appear in 8,505 images from PASCAL VOC images. Additionally, 22,394 images from ImageNet are annotated.
For every object instances, pose of the 3D object which aligned with the images are annotated in Euler angles along with ground truth bounding boxes and most similar 3D CAD models.
There are 12 general categories and 79 unique CAD models. 

%% file: sec/exp.tex
\section{Experiments}
To evaluate the effectiveness of our approach, we conduct experiments on the benchmark dataset Pascal3D+~\cite{xiang2014beyond}.
We first provide the qualitative results and quantitative comparisons with the state-of-the-art methods for both 3D shape retrieval and pose estimation tasks.
Then, we evaluate the importance of each main component of our approach.
Finally, we perform an error analysis to discuss the failure cases.

\subsection{Comparison to the State-of-the-Art}
\begin{figure*} 
\centering
\renewcommand*{\arraystretch}{0}
\begin{tabular}{*{3}{@{}c}@{}}
\includegraphics[scale=0.27]{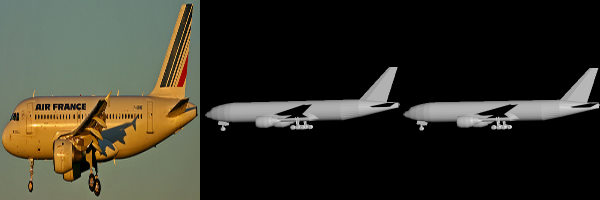} & 
\includegraphics[scale=0.27]{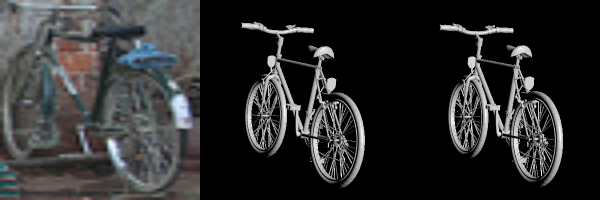} & 
\includegraphics[scale=0.27]{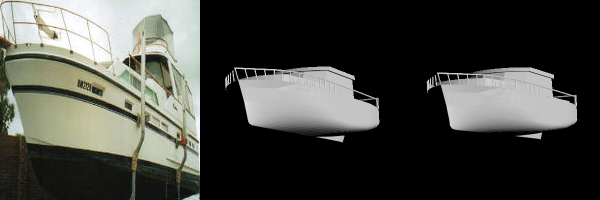} \\
\includegraphics[scale=0.27]{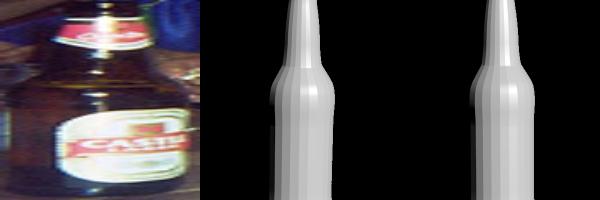} & 
\includegraphics[scale=0.27]{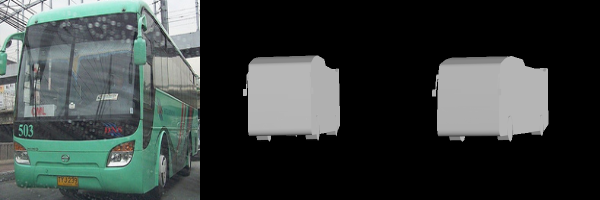} & 
\includegraphics[scale=0.27]{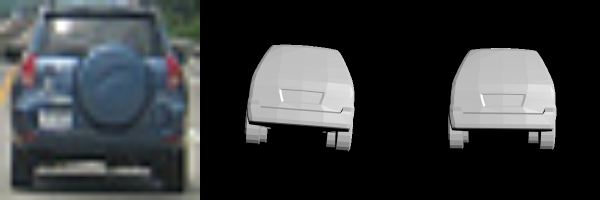} \\
\includegraphics[scale=0.27]{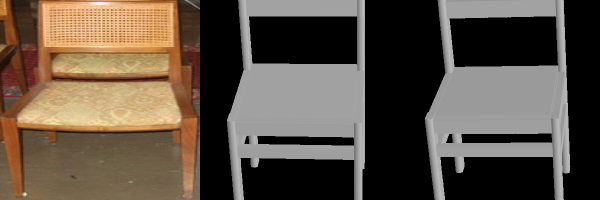} &  
\includegraphics[scale=0.27]{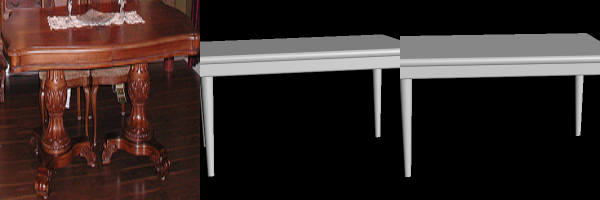} &  
\includegraphics[scale=0.27]{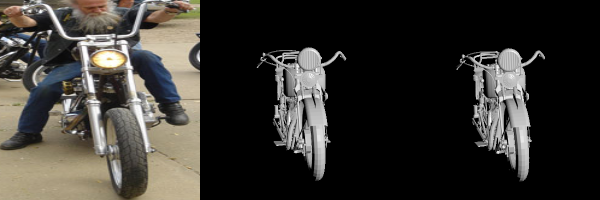} \\
\includegraphics[scale=0.27]{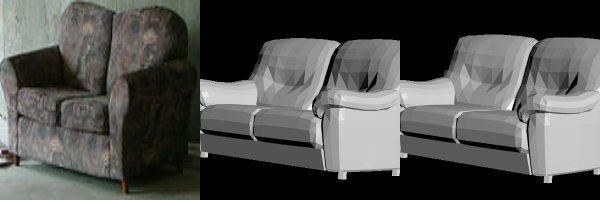}&  
\includegraphics[scale=0.27]{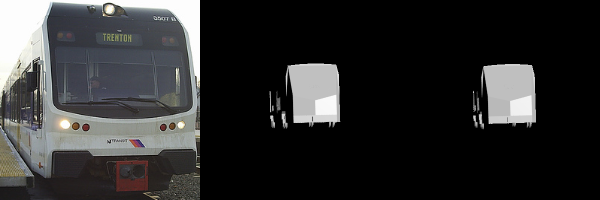}&  
\includegraphics[scale=0.27]{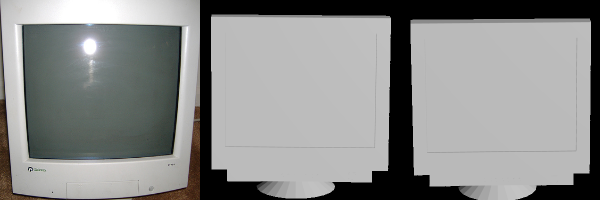} \\
\end{tabular}
\vspace{0.2cm}
\caption{Qualitative results for the pose estimation and 3d model retrieval on Pascal3D+ dataset. We show example results for all 12 categories. For each instance, the first column is the original image, the second column is image rendered using ground truth pose and model and the last column is rendered our predicted pose and model.}
\label{fig:good_results}
\end{figure*}

Following many existing methods~\cite{grabner20183d, su2015render, tulsiani2015viewpoints}, we use the images cropped with the ground truth object bounding boxes as input to our method.
%
To evaluate our method qualitatively, we render the retrieved 3D CAD model in the estimated pose.
These qualitative results with comparison to the ground truth are shown in Figure.~\ref{fig:good_results}, where one example is shown for each category.
In the following, we provide the quantitative comparisons on 3D pose estimation and 3D shape retrieval tasks.

\noindent \textbf{3D Pose Estimation}. For 3D pose estimation task, We provide a quantitative comparison against the existing methods on two metrics \textit{MedErr} and \textit{Acc$\frac{\pi}{6}$}.
\textit{MedErr} is a robust measure of pose prediction accuracy by considering the median of pose errors on all instances quantified as the geodesic distance between ground truth rotation and the predicted rotation. 
Geodesic distance between two rotations is given by Eq. \ref{eq:gd_v1}.
\textit{Acc$\frac{\pi}{6}$} is the percentage of the instances, for which the pose errors are smaller than $30^\circ$.
Following the standard evaluation protocol, we exclude truncated and occluded objects from the test dataset.

Our results are shown in Table. \ref{table:pose_comparison}.
Note that all the compared methods except for~\cite{grabner20183d,zhou2018starmap} are category-specific.
Their pose prediction network is composed of a collection of networks tailored for each category.
For Pascal3D+, there will be 12 pose prediction networks for each 12 category.
The reason for this is to exploit the biases in view point angle in the training data for a specific category.
In contrast, our pose estimation is category-agnostic by using a single sub-network for all categories, which is more scalable to datasets with a large amount of object categories.
In addition, their reported numbers are achieved by using ground truth category labels to select the specific pose networks, which is impractical in real world scenarios.
Therefore, for fair comparisons, we split Table.~\ref{table:pose_comparison} to two groups.
We can see that, our approach significantly outperforms the state-of-the-art method of~\cite{grabner20183d} in the category-agnostic setting.
Our method even outperforms all the category-specific methods in \textit{MedErr} and has comparable performance with \cite{grabner20183d} for \textit{Acc$\frac{\pi}{6}$} metric, although we do not use any ground truth labels.
The current state-of-the-art on this dataset is achieved by \cite{mahendran2018mixed}.
However, it is achieved by utilizing a vast amount of rendered data provided by \cite{su2015render}.
If we compare under the same setting, our approach significantly outperforms their method, achieving \textit{MedErr} (10.3) and \textit{Acc$\frac{\pi}{6}$} (0.8358), as compared to their \textit{MedErr} (14.3) and \textit{Acc$\frac{\pi}{6}$} (0.7506).
In addition, our category-agnostic results are even comparable to the category-specific results of \cite{grabner20183d}, outperforming their \textit{MedErr} (10.9) and almost equivalent \textit{Acc$\frac{\pi}{6}$} (0.8392).
This supports our argument that disentanglement is effective in removing categorical biases.
%

\input{sec/tables/pose_comparison.tex}
\input{sec/tables/retrieval_comparison.tex}

\noindent \textbf{3D Shape Retrieval}. We evaluate 3D shape retrieval performance of our approach using the Top-1-Acc metric and compare against~\cite{grabner20183d}, which reported the state-of-the-art 3D shape retrieval results on Pascal3D+ dataset. 
This task is challenging because some models in Pascal3D+ are quite similar to each other.
The results for 3D shape retrieval are given in Table \ref{table:retrieval_compare}.
Our approach achieved a mean Top-1-Acc of 0.592, which outperforms the state-of-the-art method of~\cite{grabner20183d} (0.460) by a large margin.
A disadvantage of their method is that its retrieval performance depends on not only the retrieval method, but also the pose estimation, since their retrieval is based on depth images rendered with the estimated pose.
Therefore, an error in pose estimation will drastically exaggerate the errors in shape retrieval.
In contrast, our approach explicitly disentangles the shape and pose embedding, and thus each embedding representation is invariant to the other.
Note that our accuracy is even higher than the highest accuracy reported in~\cite{grabner20183d} (0.4967), which is achieved by using the ground truth poses, and thus can be considered to be the theoretical upper bound of their method.

\subsection{Ablation Study}

In order to evaluate the importance of the disentangled embedding learning, we conduct experiments with two baseline methods:
1) \textbf{Ours w/o embedding}, where we only train the \textit{regression network} purely on image input in a single stage pipeline.
2) \textbf{Ours w/o disentangling}, where instead of having two separate embedding vectors of dimension 512 for both pose and shape, we use a single embedding vector of dimension 1024.
The results of the experiments are presented in Table.~\ref{table:retrieval_compare} and Table.~\ref{table:pose_ablation}.
In the following, we discuss the results of these experiments for both shape retrieval and pose estimation tasks.
\subsubsection{3D Shape Retrieval}
From Table.~\ref{table:retrieval_compare}, comparing \textbf{Ours w/o embedding} against \textbf{Ours w/o disentangling}, we can observe two stages of improvement.
First is the benefit of using 3D information free from distracting image-dependant factors in the setting \textbf{Ours w/o disentangling}.
This results in 0.02 improvement.
Second, our disentangled approach further enhances performance by another 0.02 by removing pose information from the shape embedding.
This indicates that the pose variation is also distracting for the 3D shape retrieval, which has to be removed for improved performance.

\subsubsection{3D Pose Estimation}

\noindent \textbf{Improvement by Disentangled Embedding}. From Table.~\ref{table:pose_ablation}, we can see that our disentangled embedding learning is able to significantly improve on the baseline for the pose estimation task (\textbf{Ours w/o embedding} $32$ bins  vs. \textbf{Ours}).
When comparing \textbf{Ours w/o embedding} against \textbf{Ours w/o disentangling}, we can see that the pose estimation performance degrades significantly if we use a common embedding for shape and pose to guide our \textit{regression network}.
Our interpretation of this is as follows:
The random poses applied on the 3D data is sampled from a uniform distribution.
However, the pose distribution for specific category is highly biased.
Therefore, if we do not disentangle the pose and shape to make the pose invariant to the categories, the distribution of our 3D training data does not resemble the distribution of the image data.
In contrast, the distribution of the category-independent pose embedding is closer to a uniform distribution.

\noindent \textbf{Number of Rotation Bins}. We also study the effect of different numbers of rotation bins on the pose estimation task and present results in Table.~\ref{table:pose_ablation}.
The experiments are run for 100 epochs and the best result is reported.
Using one bin is equivalent to direct regression and our result is comparable to the results presented in \cite{mahendran20173d}, which also performs direct regression.
However, we note that direct regression can achieve similar performance if run for a sufficiently long period of time.
Our experiments show that direct regression may need as much as around 500 epochs to reach similar performance in \textit{Acc}$\frac{\pi}{6}$ to $32$ bins configuration.
However, in terms of \textit{MedErr} metric, it is always consistently lower by about $1^\circ$.
We conclude that the bin and delta technique helps in two aspect.
First, classification results in a stronger supervisory signal allowing for faster convergence.
Second, delta regression is helpful in achieving more precise prediction due to the smaller range each delta sub-network has to cover.
$32$ bins setting corresponds to approximately $57^\circ$ geodesic distance between the nearest rotation for any rotation bin.
Although $72$ bins setting is slightly better, we did not use it as it performs similarly to the $32$ bins setting and $32$ bins require less memory and time for training.
The degradation which occurs at $576$ bins can be understood as an over-fitting problem.
As the number of delta networks increases, the amount of training samples seen by each network drastically decreases.

\noindent \textbf{Alternate Representation of Rotations}. We have performed evaluation of multiple rotation representations, such as axis-angles, quaternions and direct regression to rotation matrices.
The interested reader can refer to the supplementary material for the experiments.

\noindent \textbf{Visualization of Disentangled Embeddings}. We also provide visualizations of both our shape and pose embeddings in the supplmentary material, where one can clearly observe the meaningful embeddings learnt for both and shape.

\input{sec/tables/pose_ablation.tex}
\subsection{Error Analysis}

\input{sec/tables/error_analysis.tex}

We now provide a detailed analysis of our failure cases.
Both shape retrieval and pose estimation failure cases share some of the common causes: 1) blurry or small object instances, 2) ambiguity of the ground truth label and 3) truncated or occluded objects.
We provide an analysis of our failure modes in terms of object characteristics in Table. \ref{table:object_dependent_error}. 
We define 'Large Objects' to be the top one third of instances sorted by bounding box size and 'Small Objects' to be the bottom one third.
We also show the results for truncated and occluded objects for pose estimation task even though they are excluded from evaluation shown in Table. \ref{table:pose_comparison}.
We can see that both 3D shape retrieval and pose estimation has similar levels of degradation across different object characteristics and truncated objects present the most difficulty for both tasks.

\begin{figure}[h]
\centering
\renewcommand*{\arraystretch}{0}
\begin{tabular}{*{2}{@{}c}@{}}
\includegraphics[scale=0.2]{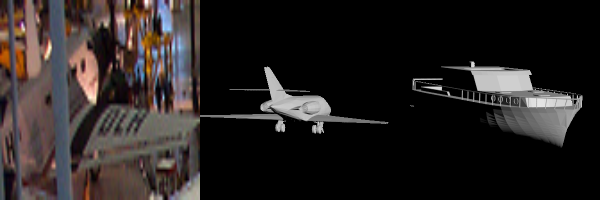}    & 
\includegraphics[scale=0.2]{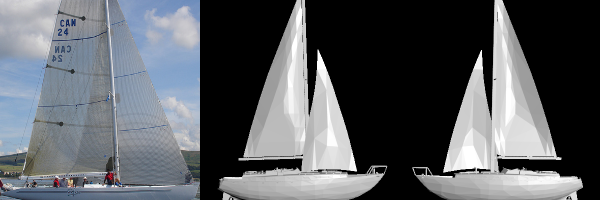}   \\
\includegraphics[scale=0.2]{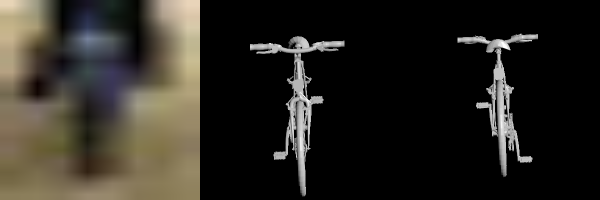}   & 
\includegraphics[scale=0.2]{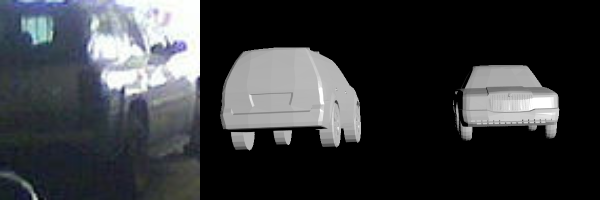}  \\
\includegraphics[scale=0.2]{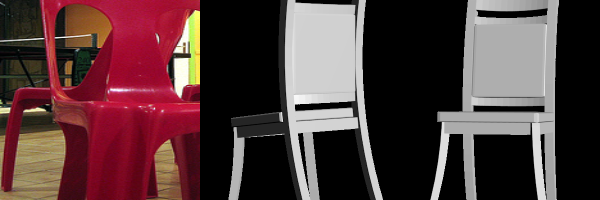}    &  
\includegraphics[scale=0.2]{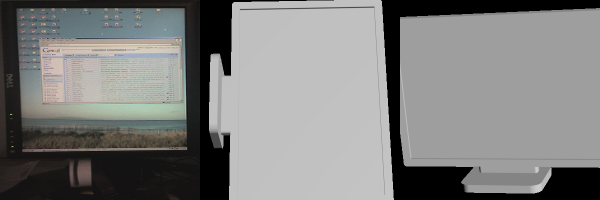}    
\end{tabular}
\vspace{0.2cm}
\caption{Example failure cases for the pose estimation task. We render the figures using predicted pose and 3D shape. Therefore, it maybe possible that both pose and retrieval are wrong in the examples. However, all the examples here have incorrect pose. Unlike the evaluation where we exclude both truncated and occluded samples, we included them in this visualization.}
\label{fig:bad_pose}

\end{figure}

\begin{figure}[h]

\centering
\renewcommand*{\arraystretch}{0}
\begin{tabular}{*{2}{@{}c}@{}}
\includegraphics[scale=0.2]{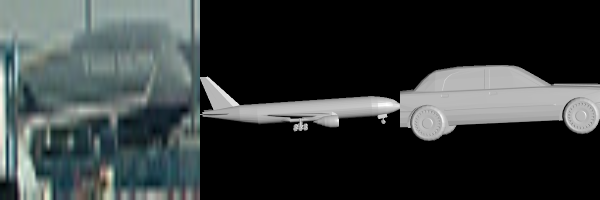}    & 
\includegraphics[scale=0.2]{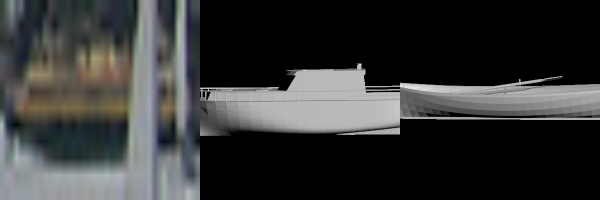}   \\
\includegraphics[scale=0.2]{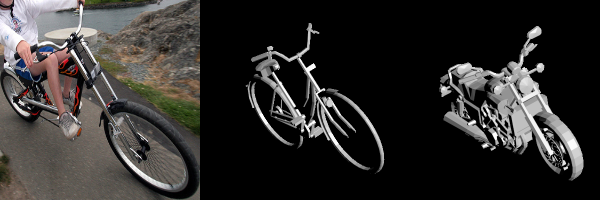}   & 
\includegraphics[scale=0.2]{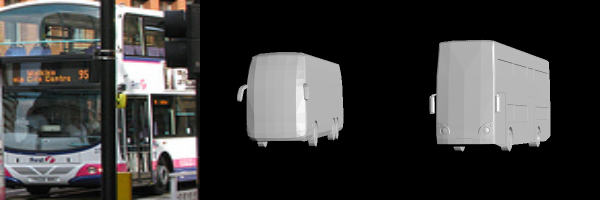}  \\
\includegraphics[scale=0.2]{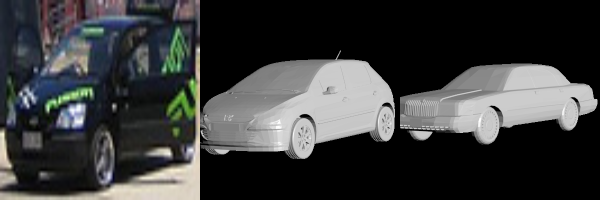}    &  
\includegraphics[scale=0.2]{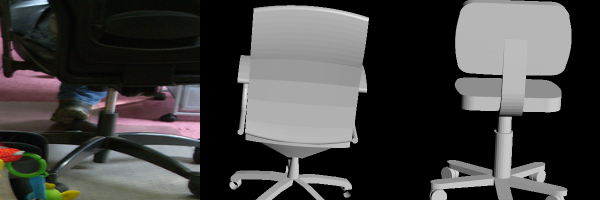}    
\end{tabular}
\vspace{0.2cm}
\caption{Example failure cases for the 3D model retrieval task. Similar to Figure. 
\ref{fig:bad_pose}, we render the figures using predicted pose and 3D shape. All the examples in this figure have incorrect shape.}
\label{fig:bad_retr}
\end{figure}

Category specific failure modes are significantly different for shape retrieval and pose estimation.
For retrieval, we obtain the lowest accuracy on chairs and sofas.
We attribute this mainly to large intra-class variation among object instances in the images and the fact that one can not find the exact or sufficiently similar 3D objects from the database.
We also note that there are several ambiguities in the annotation.
An example can be seen in Figure. \ref{fig:bad_retr} (2nd row right) where it can be difficult to determine if the ground truth bus model is closer to the image or our retrieved model.

For pose estimation, the boat score is significantly lower than other categories.
There are large intra-class differences for the boat category.
There is also significant ambiguity in many instances of boats where it is unclear which part of the boat is front and back.
In such cases, we typically fail by confusing front and back.
Although we have observed this phenomenon across all object categories (See Figure. \ref{fig:bad_pose} second row left), this problem is particularly frequent for the boat category where more than 40\% of the errors are greater than $140^\circ$, which can be interpreted as almost $180^\circ$ flip along the axis of gravity.
Other causes of error are similar to retrieval, where blurry images and ambiguity in annotation lead to errors.
Failure cases for pose estimation are shown in Figure. \ref{fig:bad_pose}.

%% file: sec/tables/pose_comparison.tex
\begin{table*}[ht]
\centering
\caption{Pose estimation comparison on Pascal3D+. Best results are highlighted in bold separately for category-specific and category-agnostic settings. * indicates the method is trained using additional rendered data from ShapeNet. 
\cite{zhou2018starmap} has provided an incorrect mean for \textit{Acc$\frac{\pi}{6}$} and \textit{MedErr} in their work.
We provide their corrected results by averaging the
\textit{Acc$\frac{\pi}{6}$} and \textit{MedErr} across 12 categories. 
}

\begin{adjustbox}{width=\textwidth}
\begin{tabular}{lcccccccccccc|c}
\Xhline{3\arrayrulewidth}
\multicolumn{14}{c}{\textbf{category-specific}}
\\
\cline{2-14}
 &   aero & bike &  boat &  bottle & bus & car & chair & dtable & mbike & sofa & train & tv & Mean  \\
\hline
$\downarrow$\textit{MedErr} (\cite{tulsiani2015viewpoints}) & 13.8 & 17.7 & 21.3 & 12.9 & 5.8 & 9.1 & 14.8 & 15.2 & 14.7 & 13.7 & 8.7 & 15.4 & 13.6 \\
$\downarrow$\textit{MedErr} (\cite{mousavian20173d})  &  13.6 & \textbf{12.5} &  22.8 & 8.3 & 3.1 & 5.8 & 11.9 & 12.5 & 12.3 & 12.8 & 6.3 & 11.9 &  11.1 \\
$\downarrow$\textit{MedErr} (\cite{su2015render}) *  &   15.4 &  14.8 &  25.6 &  9.3  & 3.6 &  6.0 &  9.7 &  \textbf{10.8}  & 16.7 &  \textbf{9.5} &  6.1  & 12.6  & 11.7 \\
$\downarrow$\textit{MedErr} (\cite{grabner20183d})  &   10.0 &  15.6 &  \textbf{19.1} &  8.6  & 3.3 &  \textbf{5.1}  & 13.7 &  11.8  & \textbf{12.2} &  13.5 &  6.7 &  \textbf{11.0}  & 10.9 \\
$\downarrow$\textit{MedErr} (\cite{mahendran2018mixed})  &  12.2 &  21.9 &  27.0  &  8.9  & \textbf{2.8}  & 5.3 & 14.6 &  25.3  & 17.5 &  16.7 &  6.1 &  13.4 & 14.3 \\
$\downarrow$\textit{MedErr} (\cite{mahendran2018mixed})*  &  \textbf{8.5} &  14.8 &  20.5  &  \textbf{7.0}  & 3.1  & \textbf{5.1} & \textbf{9.3} &  11.3  & 14.2 &  10.2 &  5.6 &  11.7  & \textbf{10.1} \\
\hline
$\uparrow$\textit{Acc$\frac{\pi}{6}$} (\cite{tulsiani2015viewpoints})  & 0.81 & 0.77 & 0.59 & 0.93 & \textbf{0.98} & 0.89 & 0.80 & 0.62 & 0.88 & 0.82 & 0.80 & 0.80 & 0.8075 \\
$\uparrow$\textit{Acc$\frac{\pi}{6}$} (\cite{mousavian20173d}) &   0.78 &  \textbf{0.83} &  0.57  & 0.93 &  0.94 &  0.90 &  0.80 &  0.68 &  0.86 &  0.82 &  0.82 &  0.85 &  0.8103 \\
$\uparrow$\textit{Acc$\frac{\pi}{6}$} (\cite{su2015render}) * &  0.74 &  \textbf{0.83} &  0.52  & 0.91 &  0.91 &  0.88 &  0.86 &  \textbf{0.73} &  0.78 &  0.90 &  \textbf{0.86}  & \textbf{0.92}  & 0.8200 \\
$\uparrow$\textit{Acc$\frac{\pi}{6}$} (\cite{grabner20183d}) &   0.83  & 0.82  & \textbf{0.64}  & 0.95 &  0.97 &  0.94 &  0.80 &  0.71 &  0.88 &  0.87 &  0.80 &  0.86 &  0.8392 \\
$\uparrow$\textit{Acc$\frac{\pi}{6}$} (\cite{mahendran2018mixed})   &  0.77  & 0.63  & 0.54 & 0.94 &  0.97 &  0.90 &   0.76 &  0.54 &  0.71 &  0.66 & 0.79 &  0.80 &  0.7506 \\
$\uparrow$\textit{Acc$\frac{\pi}{6}$} (\cite{mahendran2018mixed})*   &   \textbf{0.87}  & 0.81  & \textbf{0.64}  & \textbf{0.96} &  0.97 &  \textbf{0.95} &  \textbf{0.92} &  0.67 &  0.85 &  \textbf{0.97} &  0.82 &  0.88 &  \textbf{0.8588} \\
\Xhline{3\arrayrulewidth}
\multicolumn{14}{c}{\textbf{category-agnostic}}
\\
\cline{2-14}
 & aero & bike &  boat &  bottle & bus & car & chair & dtable & mbike & sofa & train & tv & Mean  \\
\hline
$\downarrow$\textit{MedErr} (\cite{grabner20183d})  &  10.9  &  \textbf{12.2} &  23.4 &  9.3  & 3.4 &  5.2  & 15.9 &  16.2 &  \textbf{12.2} &  11.6 & 6.3 &  \textbf{11.2} &  11.5\\
$\downarrow$\textit{MedErr} (\cite{zhou2018starmap})  & 10.1 &  14.5 &  30.0 &  9.1  & 3.1 &  6.5  & \textbf{11.0} &  23.7 &  14.1 &  11.1 & 7.4 &  13.0 &  12.8 \\
$\downarrow$\textit{MedErr} (\textbf{Ours})  & \textbf{10.0} & 13.8 & \textbf{21.1} & \textbf{7.5} & \textbf{2.8} & \textbf{4.8} & 11.4 & \textbf{10.5} & 12.9 & \textbf{9.9} & \textbf{5.4} & 13.2 & \textbf{10.3}  \\
\hline
$\uparrow$\textit{Acc$\frac{\pi}{6}$} (\cite{grabner20183d})  &  0.80  & 0.82 &  0.57 &  0.90 &  \textbf{0.97}  & \textbf{0.94} &  0.72  & 0.67 &  \textbf{0.90} &  0.80 &  0.82 &  \textbf{0.85} &  0.8133\\
$\uparrow$\textit{Acc$\frac{\pi}{6}$} (\cite{zhou2018starmap})  &  \textbf{0.82} & \textbf{0.86} &  0.50 &  0.92 &  \textbf{0.97}  & 0.92 &  \textbf{0.79}  & 0.62 &  0.88 &  \textbf{0.92} &  0.77 &  0.83 &  0.8167 \\
$\uparrow$\textit{Acc$\frac{\pi}{6}$} (\textbf{Ours})  & \textbf{0.82} & 0.84 & \textbf{0.58} & \textbf{0.93} & \textbf{0.97} & 0.91 & \textbf{0.79} & \textbf{0.76} & 0.86 & 0.90 & \textbf{0.83} & 0.83 & \textbf{0.8358}  \\
%

\Xhline{3\arrayrulewidth}
\end{tabular}
\end{adjustbox}
\label{table:pose_comparison}
\end{table*}

%% file: sec/tables/retrieval_comparison.tex

\begin{table*}[ht]
\centering
\caption{3D model retrieval accuracy using ground truth detections on Pascal3D+ in terms of Top-1-Acc.}
\begin{adjustbox}{width=1\textwidth}
\begin{tabular}{llllllllllllll}
\hline
Method & aero & bike & boat & bottle & bus  & car  & chair & table & mbike & sofa & train & tv   & mean     \\
\hline
\cite{grabner20183d} & 0.48 & 0.31 & 0.60 & 0.41   & \textbf{0.78} & 0.41 & 0.29  & 0.19  & 0.43  & 0.36  & \textbf{0.65}  & 0.61 & 0.460 \\
\hline
\textbf{Ours} w/o embedding & \textbf{0.66} & 0.50 & 0.76 & 0.51 & 0.72 & 0.51 & 0.37 & 0.52 & 0.63 & 0.33 & 0.50 & 0.66 & 0.557 \\
\textbf{Ours} w/o disentangling                                             & 0.70 & 0.55 & 0.75 & 0.54 & 0.73 & 0.54 & 0.32 & 0.52 & 0.62 & 0.51 & 0.54 & 0.59 & 0.576  \\

\textbf{Ours}  &    0.64 & \textbf{0.55} & \textbf{0.77} & \textbf{0.57} & 0.75 & \textbf{0.51} & \textbf{0.39} & \textbf{0.67} & \textbf{0.63} & \textbf{0.38} & 0.53 & \textbf{0.72} & \textbf{0.592}   \\
\hline
\end{tabular}
\label{table:retrieval_compare}
\end{adjustbox}
\end{table*}

%% file: sec/tables/pose_ablation.tex
\begin{table}[ht]
\centering
\caption{Ablation study on pose estimation performance. } 

\begin{adjustbox}{width=0.5\textwidth}

\begin{tabular}{llll}
\Xhline{3\arrayrulewidth}
Method & $\downarrow$\textit{MedErr} & $\uparrow$\textit{Acc}$\frac{\pi}{6}$ \\
\hline
Grabner et. al.\cite{grabner20183d} & 11.5 & 0.8133  \\
\hline
\textbf{Ours} w/o embedding ($1$ bin)   & 14.1 &  0.7787 \\
\textbf{Ours} w/o embedding ($8$ bins)  & 12.5  & 0.8043  \\
\textbf{Ours} w/o embedding ($32$ bins)  & 11.0 & 0.8130 \\
\textbf{Ours} w/o embedding ($72$ bins)  & 11.0  & 0.8104 \\
\textbf{Ours} w/o embedding ($576$ bins)  & 12.8 & 0.7246 \\
\hline
\textbf{Ours} w/o disentangling  ($32$ bins)     & 12.1 & 0.7951 \\
\textbf{Ours}  ($32$ bins)  & \textbf{10.3} & \textbf{0.8358} \\
\Xhline{3\arrayrulewidth}
\end{tabular}
\end{adjustbox}
\label{table:pose_ablation}
\end{table}

%% file: sec/tables/error_analysis.tex
\begin{table}[ht]
\centering
\caption{Error analysis based on object characteristics. }
\scalebox{0.95}{

\begin{tabular}{lll}
\Xhline{3\arrayrulewidth}
\multicolumn{3}{c}{\textbf{Retrieval}}\\
\hline
Setting &    & Top-1-Acc  \\
\hline
Default &  & 0.592 \\
Small Objects &  & 0.533 \\
Large Objects &   & 0.611 \\
Occluded Objects &  & 0.498 \\
Truncated Objects &  & 0.464 \\
\Xhline{3\arrayrulewidth}
\multicolumn{3}{c}{\textbf{Pose}}\\
\hline
Setting &  $\downarrow$\textit{MedErr} & $\uparrow$\textit{Acc$\frac{\pi}{6}$} \\
\hline
Default   & 10.26 & 0.8358 \\
Small Objects   & 13.17 & 0.7428 \\
Large Objects   & 9.70 & 0.8583 \\
Occluded Objects   & 18.61 &  0.6521 \\
Truncated Objects   & 21.23 & 0.6119 \\
\Xhline{3\arrayrulewidth}
\end{tabular}
}
\label{table:object_dependent_error}
\end{table}

%% file: sec/conclusion.tex
\section{Conclusion}
Joint 3D shape retrieval and pose estimation from monocular images is an important and challenging task that has a wide range of applications in robotics and augmented reality applications.
To factor out the distracting factors in the images, we learn an embedding space explicitly disentangled for shape and pose from pure 3D data, which is free from distracting factors in the images.
Our disentangled representation allows us to learn separated and more complete manifolds for pose and shape, which improves the generalization performance of our method on images of objects under unseen poses.
Our proposed method outperforms the previous state-of-the-art methods on both shape retrieval and pose estimation tasks on the challenging Pascal3D+ dataset.
As our future work, we wish to explore alternate the representation of 3D models for better discriminativity.
We are also motivated to explore further in the same direction where we map real world images to useful and task specific representation spaces.